\pdfoutput=1

\documentclass[11pt]{article}

\usepackage[]{EMNLP2022}
\usepackage{multirow}
\usepackage{times}
\usepackage{latexsym}

\usepackage[T1]{fontenc}

\usepackage[utf8]{inputenc}

\usepackage{microtype}

\usepackage{inconsolata}

\usepackage{hyperref}       
\usepackage{booktabs}       
\usepackage{amsfonts}       
\usepackage{nicefrac}       
\usepackage{microtype}      
\usepackage{xcolor}         
\usepackage{times}
\usepackage{epsfig}
\usepackage{dblfloatfix}  
\usepackage{amsmath}
\usepackage{amssymb}
\usepackage{caption}
\usepackage{tabularx}
\usepackage{booktabs}       
\usepackage{amsfonts}       
\usepackage{nicefrac}       
\usepackage{microtype}      
\usepackage{mathtools}

\usepackage{graphicx} 
\usepackage{arydshln}
\usepackage{xspace}
\usepackage{bm}

\usepackage{times}  
\usepackage{helvet}  
\usepackage{courier}  
\usepackage{url}  
\usepackage{float} 
\newcommand{\mathbbm}[1]{\text{\usefont{U}{bbm}{m}{n}#1}} 

\title{Entity-Focused Dense Passage Retrieval for \\  Outside-Knowledge Visual Question Answering}

\author{Jialin Wu\\
  Department of Computer Science \\
  The University of Texas at Austin \\
  \texttt{jialinwu@utexas.edu} \\\And
  Raymond J. Mooney \\
  Department of Computer Science \\
  The University of Texas at Austin \\
  \texttt{mooney@cs.utexas.edu} \\}

\begin{document}
\maketitle
\begin{abstract}
Most Outside-Knowledge Visual Question Answering (OK-VQA) systems employ a two-stage framework that first retrieves external knowledge given the visual question and then predicts the answer based on the retrieved content. However, the retrieved knowledge is often inadequate. Retrievals are frequently too general and fail to cover specific knowledge needed to answer the question. Also, the naturally available supervision (whether the passage contains the correct answer) is weak and does not guarantee question relevancy. To address these issues, we propose an \textbf{En}tity-\textbf{Fo}cused \textbf{Re}trieval (EnFoRe) model that provides stronger supervision during training and recognizes question-relevant entities to help retrieve more specific knowledge. Experiments show that our EnFoRe model achieves superior retrieval performance on OK-VQA, the currently largest outside-knowledge VQA dataset. We also combine the retrieved  knowledge with state-of-the-art VQA models, and achieve a new state-of-the-art performance on OK-VQA. 
\end{abstract}

\section{Introduction}
Passage retrieval under a multi-modal setting is a critical prerequisite for applications such as outside-knowledge visual question answering (OK-VQA) \cite{marino2019ok}, which requires effectively utilizing knowledge external to the image. Recently, dense passage retrievers with deep semantic representations powered by large transformer models have shown superior performance to traditional sparse retrievers such as BM25 \cite{robertson2009probabilistic} and TF-IDF under both textual \cite{karpukhin2020dense,chen2021salient,lewis2021boosted} and multi-modal settings \cite{luo2021weakly,qu2021passage,gui2021kat}. 

In this work, we investigate two main drawbacks of recent dense retrievers \cite{karpukhin2020dense,chen2021salient,lewis2021boosted,luo2021weakly,qu2021passage,gui2021kat}, which are typically trained to produce similar representations for input queries and passages containing ground-truth answers. 

\begin{figure}[t]
    \centering
    \includegraphics[clip, trim=1cm 11cm 37cm 0cm, width=0.9\columnwidth]{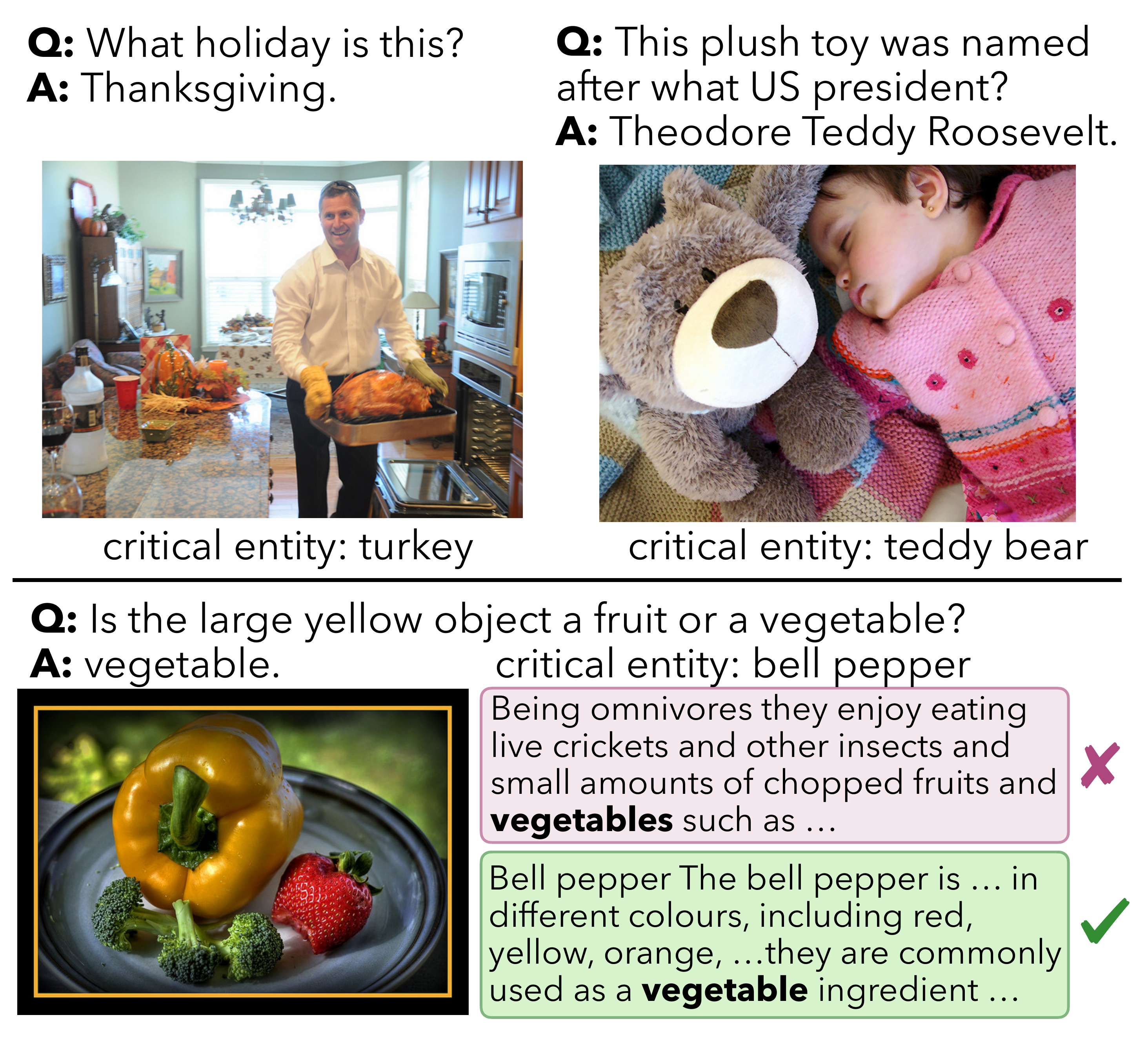}
    \caption{Top: Examples of critical entities upon which retrieval models should focus; Bottom: Example of improved passage retrieval using critical entities. }
    \label{fig:teaser}
\end{figure}

First, as most retrieval models encode the query and passages as a whole, they fail to explicitly discover entities critical to answering the question \cite{chen2021salient}. This frequently leads to retrieving overly-general knowledge lacking a specific focus. Ideally, a retrieval model should identify the critical entities for the query and then retrieve question-relevant knowledge specifically about them. For example, as shown in the top half of Figure \ref{fig:teaser}, retrieval models should realize that the entities ``turkey'' and ``teddy bear'' are critical.

Second, on the supervision side, the positive signals are often passages containing the right answers with top sparse-retrieval scores such as BM 25 \cite{robertson2009probabilistic} and TF-IDF. However, this criterion is inadequate to guarantee question relevancy, since good positive passages should reveal facts that actually support the correct answer using the critical entities depicted in the image. For example, as shown in the bottom of Figure \ref{fig:teaser}, both passages mention the correct answer ``vegetable'' but only the second one which focuses on the critical entity ``bell pepper'' is question-relevant.

In order to address these shortcomings, we propose an \textbf{En}tity-\textbf{Fo}cused \textbf{Re}trieval (EnFoRe) model that improves the quality of the positive passages for stronger supervision. EnFoRe automatically identifies critical entities for the question and then retrieves knowledge focused on them. We focus on entities that improve a sparse retriever's performance if emphasized during retrieval as critical entities. We use the top passages containing {\it both} critical entities and the correct answer as positive supervision. Then, our EnFoRe model learns two scores to indicate (1) the importance of each entity given the question and the image and (2) a score that measured how well each entity fits the context of each candidate passage. 

We evaluate EnFoRe on OK-VQA \cite{marino2019ok}, currently the largest knowledge-based VQA dataset. Our approach achieves state-of-the-art (SOTA) knowledge retrieval results, indicating the effectiveness of explicitly recognizing key entities during retrieval. We also combine this retreived  knowledge with SOTA OK-VQA models and achieve a new SOTA OK-VQA performance. Our code is available at \url{https://github.com/jialinwu17/EnFoRe.git}.
\section{Background and Related Work}

\subsection{OK-VQA}
Visual Question Answering (VQA) has witnessed remarkable progress over the past few years, in terms of both the scope of the questions \cite{antol2015vqa,hudson2019gqa,wang2018fvqa,gurari2018vizwiz,singh2019towards}, and the sophistication of the model design \cite{antol2015vqa,lu2016hierarchical,anderson2017bottom,kim2018bilinear,kim-etal-2020-conan,wu2019generating,wu2019self,jiang2018pythia,lu2019vilbert,nguyen2020movie}. There is a recent trend towards outside knowledge visual question answering (OK-VQA) \cite{marino2019ok}, where open domain external knowledge outside the image is necessary. Most OK-VQA models \cite{marino2019ok,garderes2020conceptbert,zhu2020mucko,li2020boosting,narasimhan2018out,marino2020krisp,wu2021multi,gui2021kat} incorporate a retriever-reader framework that first retrieves textual knowledge relevant to the question and image and then ``reads'' this text to predicts the answer. As an online free encyclopedia, Wikipedia is often used as the knowledge source for OK-VQA. While most previous works focused more on the answer prediction stage, the performance is still lacking because of the imperfect quality of the retrieved knowledge. This work focuses on knowledge retrieval and aims at retrieving question-relevant knowledge that focuses explicitly on the critical entities for the visual question.

\subsection{Passage Retrieval}

\noindent\textbf{Sparse Retrieval:}
Before the recent proliferation of transformer-based dense passage retrieval models \cite{karpukhin2020dense}, previous work mainly explored sparse retrievers, such as TF-IDF and BM25 \cite{robertson2009probabilistic}, that measure the similarity between the search query and candidate passage using weighted term matching. 
These sparse retrievers require no training signals on the relevancy of the passage and show solid baseline performances. However, exact term matching prevents them from capturing synonyms and paraphrases and understanding the semantic meanings of the query and the passages.  

\noindent\textbf{Dense Retrieval:}
To better represent semantics, dense retrievers \cite{karpukhin2020dense,chen2021salient,lewis2021boosted,lee2021phrase} extract deep representations for the query and the candidate passages using large pretrained transformer models. Most dense retrievers are trained using a contrastive objective that encourages the representation of the query to be more similar to the relevant passages than other irrelevant passages. During training, the passage with a high sparse retrieval score containing the answer is often regarded as a positive sample for the question-answering task. However, these positive passages may not fit the question's context and only serve as very weak supervision.
Moreover, the query and passages are often encoded as single vectors. Therefore most dense retrievers fail to explicitly discover and utilize critical entities for the question \cite{chen2021salient}. This often leads to overly general knowledge without a specific focus.

\subsection{Dense Passage Retrieval for VQA}
Motivated by the trend toward dense retrievers, previous work has also applied them to OK-VQA. \citet{qu2021passage} utilize Wikipedia as a knowledge source. \citet{luo2021weakly} crawl Google search results on the training set as a knowledge source. However, the weak training signals for passage retrieval become more problematic for VQA as the visual context of the question makes it more complex. Therefore, a ``positive passage'' becomes less likely to fit the visual context and actually provide suitable supervision. In order to better incorporate visual content, \citet{gui2021kat} adopt an image-based knowledge retriever that employs the CLIP model \cite{radford2021learning} pretrained on large-scale multi-modal pairs as the backbone. However, question relevancy is not considered, so the retriever has to retrieve knowledge on every aspect of the image for different possible questions.

This work proposes an \textbf{En}tity-\textbf{Fo}cused \textbf{Re}trieval (EnFoRe) model that recognizes key entities for the visual question and retrieves question-relevant knowledge specifically focused on them. Our approach also benefits from stronger passage-retrieval supervision with the help of those key entities.

\subsection{Phrase-Based Dense Passage Retrieval}
The most relevant work to ours is phrase-based dense passage retrieval.  \citet{chen2021salient} employ a separate lexical model that is trained to mimic the performance of a sparse retriever that is better at matching phrases. \citet{lee2021phrase} propose DensePhrase model that extracts each possible phrase feature in the passage and only uses the most relevant phrase to measure the similarity between the query and passage. However, the training signals still come from exactly matching ground truth answers, and the phrases are parsed from the candidate passage, limiting the scope of the search. 
In contrast, our approach collects entities from many aspects of the question and image, including object recognition, attribute detection, OCR, brands, captioning, etc., building a rich unified intermediate representation. 

\section{Entity Set Construction}
Our EnFoRe model is empowered by a comprehensive set of extracted entities. \ Entities are not limited to phrases from the question and passages as in \cite{lee2021phrase}. We collect entities from the sources below.  Most entity extraction steps are independent and can execute in parallel, except for answering sub-questions, which first requires parsing the questions.  Parallelizing these steps can significantly reduce run time.

\subsection{Question-Based Entities}

\noindent\textbf{Entities from Questions:} First, the noun phrases in questions usually reveal critical entities. Following \citet{wu2021multi}, we parse the question using a constituency parser \cite{allennlp} and extract noun phrases at the leaves of the parse tree. Then, we link each phrase to the image and extract the referred object with its attributes. We use a pretrained ViLBERT model \cite{Lu_2020_CVPR} as the object linker.

\begin{figure*}[t]
    \centering
    \includegraphics[clip, trim=0cm 23.5cm 45cm 1cm, width=1.6\columnwidth]{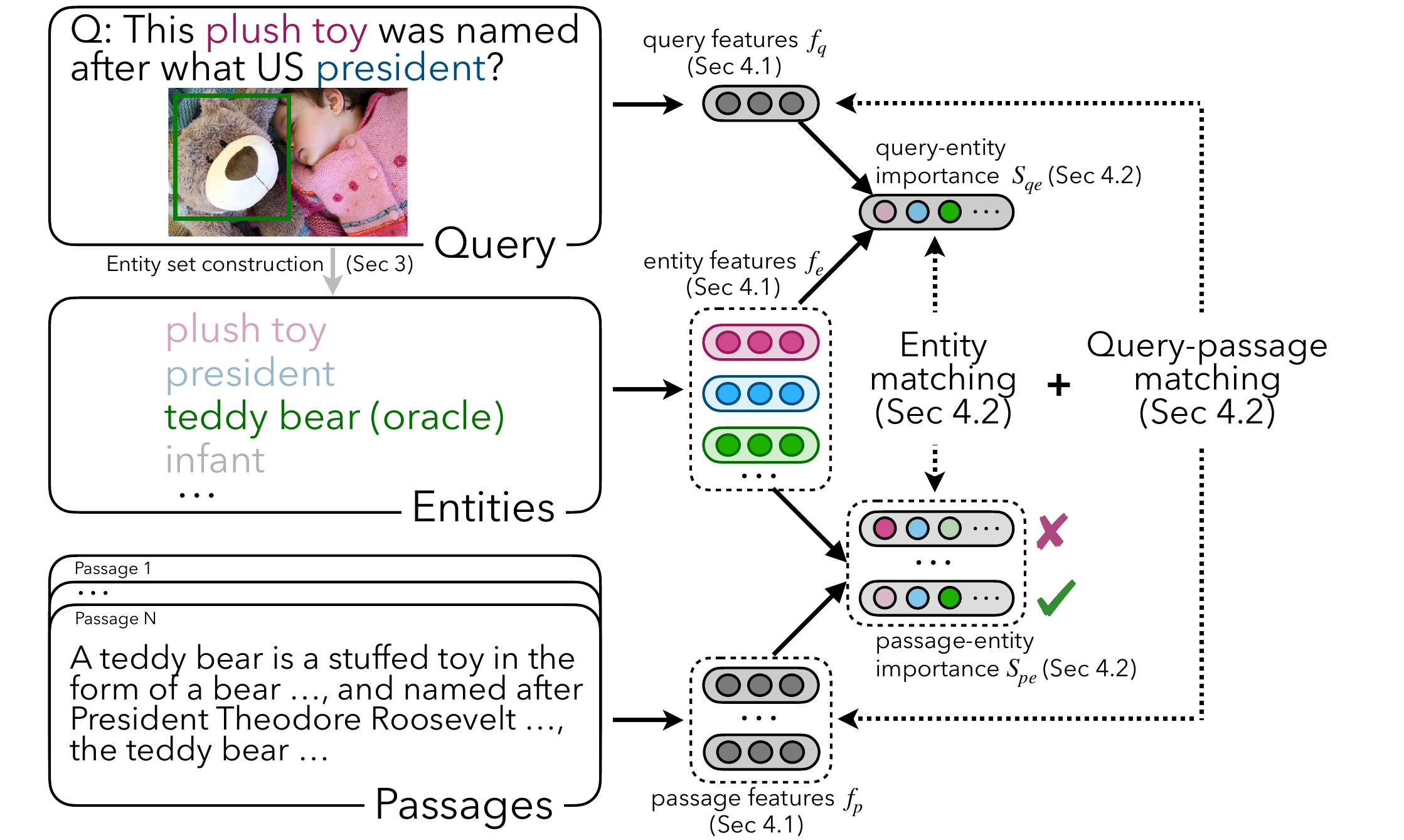}
    \caption{EnFoRe model overview. We first extract a set of entities from the query consisting of a question and an image (Sec. 3). Then, the EnFoRe model computes the features for the query, the entities, and the passages (Sec. 4.1). Query features and passage features, together with entity features, are used to compute a query-entity score and a passage-entity score to indicate the importance of the entities given the query and the passages, respectively (Sec. 4.2). These two importance scores are combined to produce an entity-matching score, and the features of the query and the passages are used to predict a query-passage matching score. }
    \label{fig:model}
\end{figure*}

\noindent\textbf{Entities from Sub-Questions:} OK-VQA often requires systems to solve visual reference problems as well as comprehend relevant outside knowledge. Therefore, we employ a general VQA model to find answers to the visual aspects of the question. In particular, we collect a set of sub-questions by appending each noun phrase in the parse tree to the common question phrases ``What is...'' and ``How is...'' When the confidence for an answer from a pretrained VilBERT model \cite{Lu_2020_CVPR} exceeds 0.5, it is added to the entity set. 
For the example in Fig.\ \ref{fig:model}, the noun phrases ``plush toy'' and ``president'' generate the sub-questions: “What is plush toy?”, “How is plush toy?”, “What is president?”, “How is president?”. The answer confidence for ``teddy bear'' exceeds 0.5 for the first question, so we include it in the entity set.

\noindent\textbf{Entities from Answer Candidates:} Standard state-of-the-art VQA models are surprisingly effective at generating a small set of promising answer candidates for OK-VQA \cite{wu2021multi,wu2020improving}. Therefore, we finetune a ViLBERT model \cite{lu2019vilbert} on the OK-VQA data set and extract the top $5$ answer candidates and add them to entity set.

\subsection{Image-Based Entities}

Question-based entities are high precision and narrow down the search space for knowledge retrievers. To complement this, we also collect image-based entities to help achieve higher recall.

\noindent\textbf{Entities from Azure tagging:} Following \citet{yang2021empirical}, we use Azure OCR and brand tagging to annotate the detected objects in the images using a Mask R-CNN detector \cite{he2017mask}.

\noindent\textbf{Entities from Wikidata:} As suggested by \citet{gui2021kat}, common image and object tags can be generic with a limited vocabulary, leading to noise or irrelevant knowledge. Therefore, we also leverage recent advanced visual-semantic matching approaches, i.e. CLIP \cite{radford2021learning}, to extract image-relevant entities from Wikidata. In particular, the entities with their descriptions in Wikidata and sliding windows of the images are used as inputs. Then, at most $18$ entities with top maximum CLIP scores over these sliding windows are preserved. We follow the released code for KAT \cite{gui2021kat} and resize the image such that the size of the shorter edge is 384. The sliding window size is set to 256 with a stride of 128.

\noindent\textbf{Entities from Captions:} Captions provide a natural source of salient objects in the image, and do not suffer from the limited vocabulary of object detectors \cite{wu2018joint}. Similar to extracting entities from the question, we parse captions and extract noun phrases from the parse tree. During training, we use the human captions provided by the COCO dataset to provide richer entities, and during testing, we use generated captions from the OFA captioning model \cite{wang2022unifying}.

\subsection{Oracle Critical Entity Detection}
\label{sec:oracle_entity}
Given the comprehensive set of entities $\mathcal{E}$  covering different aspects of the question and image, we introduce an approach to  automatically find critical entities and passages containing them. Then, those entities and passages are used during training to provide more substantial supervision. The intuition is that a good passage that fits the visual question's context should mention {\it both} the key entities and the correct answer. Also, emphasizing critical  entities should improve retrieval performance.

Given a question $q$, we use BM25\footnote{ https://github.com/castorini/pyserini.git} \cite{robertson2009probabilistic} as the sparse retriever to retrieve an initial set of passages $\mathcal{P}_{init} = \{p_1, ..., p_K\}$. We calculate a baseline score $\texttt{SRR}_{init}$ for these $K$ passages using summed reciprocal ranking (\texttt{SRR}) as shown in Eq. \ref{eq:srr}. 
\begin{equation}
    \texttt{SRR}(\mathcal{P}) = \sum_{i = 1}^{K} \frac{ \mathbbm{1}[ \texttt{ans} \in p_i] }{i} \label{eq:srr}
\end{equation}
We use summed reciprocal ranking instead of reciprocal ranking since it provides more stable scores for evaluating the set of retrieved passages and does not overweight the highest ranked document.

Then, for each entity $e \in \mathcal{E}$, we retrieve another set of passages $\mathcal{P}_e$ using an entity-emphasizing query where the entity is appended to the end of the question. Note that the BM25 retriever does not take word order into account, so simply appending entities will not lead to undesired results due to the linguistic disfluency of the query. 

The final score for an entity $S(e)$ is computed as the difference between the $\texttt{SRR}$ of these two sets of retrieved passages, i.e. $S(e) = \texttt{SRR}(\mathcal{P}_e) - \texttt{SRR}(\mathcal{P}_{init})$. We regard entities with $S(e)$ over a threshold $\theta$ as critical entities, i.e. $\mathcal{E}_{oracle} = \{e \in \mathcal{E}| S(e) > \theta\}$.

\citet{qu2021passage} extract the top-$k$ passages containing the correct answer from $\mathcal{P}_{init}$ to construct the positive passage set $\mathcal{P}^+_{init}$.  As we have identified oracle entities, the passage that contains both the right answer and the oracle entity is more likely to fit in the context of the question. 
Therefore, we augmented the positive passage set to include those passages for each oracle entity, $i.e.$ $\mathcal{P}^+_{\mathcal{E}} = \bigcup_{e \in \mathcal{E}_{oracle}}( \{p^+_e \})$, where $p^+_e$ denotes the first passage that contains both the right answer and the oracle entity. On average, there are 3.4 new positive passages per question.  The negative passages are the same as those in \cite{qu2021passage}, and the number of training instances (positive-negative pairs) is not changed.

\section{Entity-Focused Retrieval}
\textbf{En}tity-\textbf{Fo}cused \textbf{Re}trieval (EnFoRe) automatically recognizes critical entities and retrieves question-relevant knowledge specifically focused on them. ``\texttt{proj}'' denotes a projection function that consists of an MLP layer with layer-norm as normalization.

\subsection{Encoders}
\noindent\textbf{Query encoder:} As observed by \citet{qu2021passage} and \citet{luo2021weakly}, multi-modal transformers encode questions  and  visual content better than uni-modal transformers, so we adopt LXMERT \cite{tan2019lxmert} for query encoding. In particular, we project the ``\texttt{pooled\_output}'' at the last layer from LXMERT as the feature vector $f_q \in R^d$ given the query $q$ that contains a visual question $Q$ and the set of detected objects $\mathcal{V}$ in the image as shown in Eq. \ref{eq:query_encoder}. See the LXMERT paper  for further details. 
\begin{equation}
    f_q =  \texttt{proj}(\texttt{LXMERT}(Q, \mathcal{V})) \label{eq:query_encoder}
\end{equation}\\

\noindent\textbf{Passage encoder:} Following \citet{qu2021passage}, we use BERT \cite{devlin2018bert} as the passage encoder and project the ``\texttt{[CLS]}'' representation to compute the vector features for each passage $p$.
\begin{equation}
    f_p = \texttt{proj}(\texttt{BERT}(p))
\end{equation}

\noindent\textbf{Entity encoder:}  In order to provide query context for each entity, we append the question and a generated image caption \cite{wang2022unifying} after each entity. The input to the Entity encoder is ``\texttt{[CLS] entity [SEP] question [SEP] caption}''. Similar to the passage encoder, we use BERT \cite{devlin2018bert} as the entity encoder and project the ``\texttt{[CLS]}'' representation to compute the features for each entity.
\begin{equation}
    f_e = \texttt{proj}(\texttt{BERT}(e))
\end{equation}

\subsection{Retrieval Scores }
EnFoRe aims to retrieve question-relevant knowledge that focuses on critical entities. Therefore, the similarity metric consists of two parts: a question relevancy term and an entity focus term.

\noindent\textbf{Modeling question relevancy:} We model the question relevancy term $S_{qp}$ as the inner-product of the query and passage features, i.e. $S_{qp}(q, p) =  f_q^T  f_p$. During inference, as the query and passage features are decomposable, maximum inner product search (MIPS) can be applied to efficiently retrieve top passages for the query. 

\noindent\textbf{Modeling entity focus:} The entity focus term consists of two parts, where query features are used to identify critical entities from the set of entities in Sec. 3, and passage features are used to determine whether it contains these key entities. For each entity,  we compute the query-entity score $S_{qe}(q, e)$ as the inner-product of the projected query and entity feature, i.e. $S_{qe}(q, e) = \texttt{proj}(f_q)^T\texttt{proj}(f_e)$, and we compute the passage-entity score as  $S_{pe}(p, e) = \texttt{proj}(f_p)^T\texttt{proj}(f_e)$. Then, we combine all of the entities and compute the entity-focused score $S_{qpe}$ per Eq. \ref{eq:entity_focusing}:
\begin{equation}
    S_{qpe}(q, p, \mathcal{E}) = \frac{\sum_{e \in \mathcal{E}} \sigma(S_{qe}(q, e))\times S_{pe}(p, e)}{\sum_{e \in \mathcal{E}} \sigma(S_{qe}(q, e))} \label{eq:entity_focusing}
\end{equation}
where $\sigma$ denotes the sigmoid function.
Another way to interpret Eq. \ref{eq:entity_focusing} is to treat it  as modeling the conditional distribution $\Pr(p \mid q)$ and consider the entities as hidden variables.

The final score $S(q, p)$ for the query $q$ and passage $p$ linearly combines both terms, i.e. $S(q, p) = S_{qp}(q, p) + \lambda S_{qpe}(q, p, \mathcal{E})$, where the weight $\lambda$ controls the balance between the these two terms.

\begin{table*}[t]
\centering
\small
\begin{tabular}{l|c|c||c|c}
\toprule
 Methods                            & \multicolumn{2}{c||}{Val} & \multicolumn{2}{c}{Test} \\\hline
                                  & MRR@5       & P@5       & MRR@5        & P@5       \\\hline
 BM25-Obj                           &   0.3772    &  0.2667   &  0.3686      &  0.2541  \\
BM25-Cap                           &    0.4727   &  0.3483   &  0.4622      &  0.3367   \\
BM25 w. entities                   &    0.3620   &  0.2558   &  0.3732      &  0.2620   \\
BM25 w. oracle entities            &   0.6591   &   0.4548   &  0.6401      &  0.4345     \\\hline
DPR-LXMERT    \cite{qu2021passage}  &    0.4704   &  0.3364   &  0.4526      &  0.3329  \\
EnFoRe-LXMERT                      &   \textbf{0.4881}   &  \textbf{0.3488} &  \textbf{0.4800}  &  \textbf{0.3444}\\\hline
EnFoRe-LXMERT w. oracle entities    &    0.4898   &  0.3533  &  0.4853 &  0.3451   \\\bottomrule
\end{tabular}
\caption{MRR and precision retreival results on OK-VQA. The first four rows present sparse retrieval results and the others are dense retrieval results.  }
\label{tab:retriever_result}
\end{table*}

\subsection{Training} We train our EnFoRe model with a set of training instances consisting of a query containing the visual question with an image, a positive passage, a retrieved negative passage, and the set of entities. We present more details on constructing the training data in Sec. \ref{sec:dataset}. We adopt the ``R-Neg+IB-All'' setting introduced by \citet{qu2021passage} that regards the retrieved negatives, along with all other in-batch passages, as negative samplings. Following previous work \cite{karpukhin2020dense}, we use cross-entropy loss to maximize the relevancy score $S_{qp}(q, p)$ and the entity focusing score $S_{qpe}(q, p, \mathcal{E})$ of the positive passage given the negatives identified above.  In addition, we regard the oracle entities, defined in Sec.\ \ref{sec:oracle_entity}, as positive entities and others as negative entities. We use binary cross-entropy loss to supervise the importance score $S_{qe}(q, e)$. 
We use AdamW \cite{loshchilov2017decoupled} with a learning rate of 1e-5 to train the EnFoRe model for 8 epochs where 10\% of the iterations are used to warm up the model linearly. The batch size is set to 6 per GPU, and we use 4 GPUs (Tesla V100) for each experiment. The training process takes about 45 hours for each model. We save the parameters every 5000 steps and present the best results (MRR@5) on the validation set. The hidden states size is set to 768 following \citet{qu2021passage} for fair comparison. The threshold $\theta$ for recognizing critical entities is set to 0.8. As our model consists of a BERT encoder and a LXMERT encoder, resulting in 430M parameters in total.

\noindent\textbf{Inference:} As the question relevancy term is decomposable, we again adopt MIPS to retrieve the top-80 passages. Then, we evaluate the entity focus term for each passage and use the combined score $S(q, p)$ to rerank the retrieved passages.

\section{Reader}
We employ the current state-of-the-art KAT model \cite{gui2021kat} as our VQA reader. KAT is a generation-based reader that learns to generate the answer given the retrieved knowledge. It adopts an FiD \cite{izacard2020distilling} architecture to incorporate both implicit knowledge, generated by a frozen GPT-3 model, and explicit knowledge. For implicit GPT-3 knowledge, the input format is ``\texttt{question:ques?}\texttt{candidate:cand.} \texttt{evidence:expl.}'', where the \texttt{ques}, \texttt{cand} and \texttt{expl.} denotes the question, answer and its explanation generated by the GPT-3 model \cite{brown2020language}. For the explicit knowledge, the input format is ``\texttt{question:ques?} \texttt{entity:ent.} \texttt{description:desc.}'', where \texttt{ent}, \texttt{desc} denote the retrieved entity and its description. See \cite{gui2021kat} for further details.

\begin{table*}[t]
\centering
\small
\begin{tabular}{l|ccc|ccc|c|c}
\toprule
                  & \multicolumn{3}{c|}{Image-based entities}   & \multicolumn{3}{c|}{Question-Image-based entities}                                    & MRR@5 & P@5 \\\hline
                  & Tags & Wikidata & Cap. & Ques. & Sub-Ques. & Cand. &       &     \\\hline
DPR-LXMERT    &    &       &       &                  &                  &           &    0.4526  & 0.3329   \\
EnFoRe (Backbone)    &    &       &       &                  &                  &           &    0.4632   & 0.3317    \\
EnFoRe (Image)    &  \checkmark    &  \checkmark   &  \checkmark   &                  &                         &       &    0.4688     &  0.3351   \\
EnFoRe (Question)    &  &   &     &    \checkmark     &    \checkmark     &           \checkmark   &  0.4750     &  0.3409   \\
EnFoRe (Full)     & \checkmark  &    \checkmark     &    \checkmark        &    \checkmark     &       \checkmark      &    \checkmark    &  0.4800     &  0.3444  \\\bottomrule
\end{tabular}
\caption{Ablation study on the entity sources used during re-ranking.}\label{tab:entity_source}
\end{table*}

We change the original explicit knowledge to the knowledge retrieved by our EnFoRe model. As the retrieved passage contains multiple sentences, and usually not all are relevant, we select the most relevant sentence for each passage. Specifically, following \citet{wu2021multi}, we convert the question and the candidate answers to a set of statements. Then, we decontextualize each sentence for each passage and compute the BertScore \cite{zhang2019bertscore} between the decontextualized sentences and each statement. The sentence with the highest BertScore across these statements is extracted for each passage. The input format is ``\texttt{question:ques?}\texttt{entity:ents.} \texttt{description:desc.}'', where the \texttt{ents}, \texttt{desc} denote the top-10 entities judged by the query-entity importance score $S_{qe}(q, e)$ and the extracted sentence. 

Following \citet{gui2021kat}, we perform experiments for two KAT settings: (1) ``KAT-base + EnFoRe'' setting is a single model that employs T5-base \cite{2020t5} as the backbone encoder and decoder. (2) ``KAT-full + EnFoRe'' is an ensemble model, where each model employs T5-large as the backbone encoder and decoder. As our knowledge is question-aware, we only encode the top 10 retrieved sentences in contrast to the 40 sentences in the original KAT. We adopt the same training scheme as KAT.



\section{Experimental Results}
\subsection{Dataset}
\label{sec:dataset}
We use the OK-VQA dataset\footnote{https://okvqa.allenai.org/} \cite{marino2019ok} (version 1.1), the largest open-domain English knowledge-based VQA dataset at present, to evaluate the EnFoRe model. The questions were crowd-sourced on Amazon Mechanical Turk (AMT) and are guaranteed to require external knowledge beyond the images. The dataset contains 14,031 images and 14,055 questions covering a variety of knowledge categories (i.e. 9,009 for training and 5046 for test). For knowledge retrieval, we adopt the same data configuration as \citet{qu2021passage} that evenly splits the test set of the OK-VQA dataset into a validation set and a test set, and we refer to these as RetVal and RetTest, respectively. 

Following \citet{qu2021passage}, we take the Wikipedia passage collection with 11 million passages created by previous work as our knowledge source, where each passage contains at most 384 “word pieces” with intact sentence boundaries. We extract 25 passages with the highest BM 25 scores (CombSum setting in \cite{qu2021passage}) that do not contain the correct answers
as our retrieved negative samples, and the top 5 passages that contain the correct answer as retrieved positive samples for training. In addition, we also consider the most relevant passage that contains each of the oracle entities and the correct answer as positive passages. The positive and negative passages are randomly paired up to form the training instances. During evaluation, any passages containing at least one of the correct answers are considered as gold passages.
For VQA models, we adopt the same model architecture and training scheme and only switch the external knowledge for the KAT models. Due to limits on computational resources, we adopt 10 retrieved sentences for the KAT model. The models are evaluated every 500 steps. We normalize the predictions by lowercasing, lemmatizing, and removing articles, punctuation and duplicated whitespace. We follow the standard evaluation metric recommended by the VQA challenge.\footnote{https://github.com/GT-Vision-Lab/VQA} The results for ``KAT-base + EnFoRe'' are obtained by averaging three runs with different random seeds.

\begin{table*}[t]
\centering
\small
\begin{tabular}{l|c|c}
\toprule
\textbf{Method}  & \textbf{Knowledge Resources} & \textbf{VQA Scores} \\\hline
Q-only \cite{marino2019ok}  & --- &  14.9   \\
BAN \cite{kim2018bilinear}  & --- & 25.2   \\
MUTAN \cite{ben2017mutan}  & --- &  26.4 \\
Mucko \cite{zhu2020mucko}  &  Dense Caption  &  29.2  \\ 
ConceptBert \cite{garderes2020conceptbert}  &  ConceptNet  &  33.7 \\
KRISP \cite{marino2020krisp}  & Wikipedia +  ConceptNet  &  38.9 \\  
MAVEx \cite{wu2021multi} & Wikipedia +ConceptNet + Google Image & 39.4 \\ 
RVL \cite{shevchenko2021reasoning}  &  Wikipedia + ConceptNet  &  39.0 \\
VRR \cite{luo2021weakly} & Google Search & 39.2 \\
PICa \cite{yang2021empirical} & Frozen GPT-3  &  48.0 \\ \hline
KAT-base  & Frozen GPT3 + Wikidata & (50.58) \\    
KAT-base + EnFoRe & Frozen GPT3 + Wikipedia & 51.34 (52.24)\\\hline
KAT-full  &  Frozen GPT3 + Wikidata &  (54.41) \\   
KAT-full + EnFoRe &  Frozen GPT3 + Wikipedia & 54.35 (55.23)\\
\bottomrule
\end{tabular}

\caption{EnFoRe knowledge boosts the current state-of-the-art approaches on OK-VQA. The middle column lists the external knowledge sources if any, used in each system. The additional result shown in parentheses is computed by an unofficial evaluation metric 
that takes the max over 1.0 and number of annotators agreements divided by 3.}
\label{tab:vqa_result}
\end{table*}

\subsection{Passage Retrieval Results}
We present our passage retriever results in Table \ref{tab:retriever_result}, comparing them with the current state-of-the-art systems. We adopt MRR and Precision at a cut-off of 5 as our automatic evaluation metric. The first four rows present sparse retrieval results. The BM25 approach using our oracle entities achieves an MRR@5 of 0.6401, and a precision@5 of 0.4345 on the OK-VQA RetTest set, indicating the comprehensiveness and the potential helpfulness of the extracted entities. With the help of these entities,  EnFoRe-LXMERT outperforms the previous SOTA DPR-LXMERT (with the same architecture for visual and textual embedding) by 2.74\%  MRR@5 and 1.15\% precision@5. 
We perform a student's paired t-tests with a p-value of 5\% to test the significance of our results. In particular, we found that the MRR and the precision gap between our EnFoRe (Full) model and (1) the DPR-LXMERT and (2) the EnFoRe (Backbone) are statistically significant.

\noindent\textbf{Ablation study on entity sources:}
We also performed an ablation study on entity-based re-ranking shown in Table \ref{tab:entity_source}. The EnFoRe backbone without re-ranking achieves an MRR of 0.4632, outperforming DPR \cite{qu2021passage} by 1.06\%. This indicates that using our entities during training helps the retriever build better representations. It is because (1) we add additional supervision that tells the retriever which entities are more likely to lead to the correct answers, and (2) we add additional training passages that contain both the oracle entities and the right answers. Image-based and Question-based entities help our EnFoRe model achieve MRR of 0.4688 and 0.4750, respectively. Our full model, taking advantage of both image- and question-based entities, achieves an MRR of 0.4800, showing that these two types of entities are complementary.

\begin{figure*}[t]
    \centering
    \includegraphics[clip, trim=0cm 26.2cm 21cm 0cm, width=\linewidth]{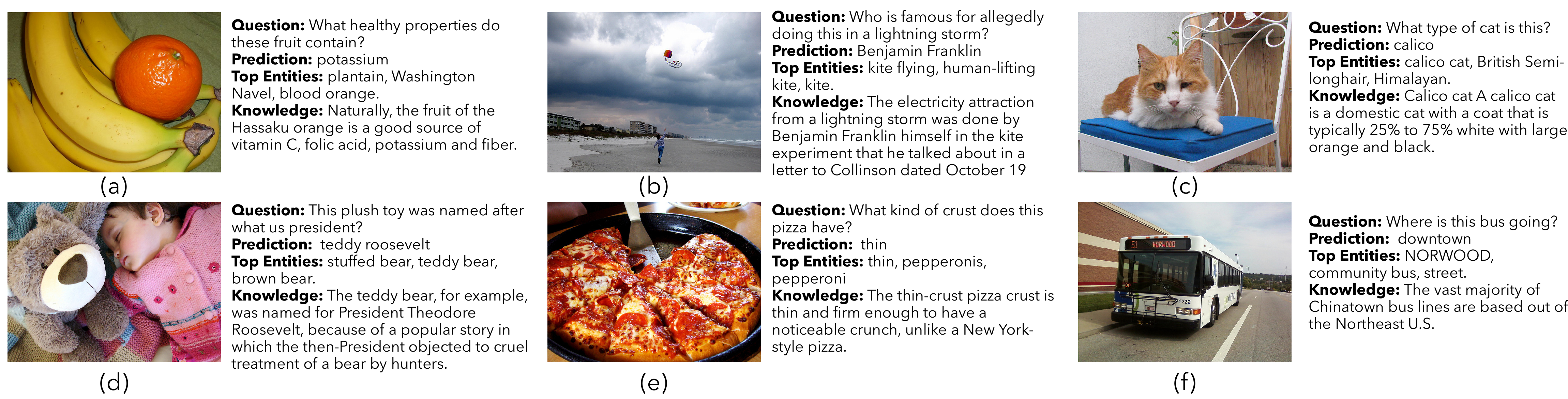}
    \caption{Qualitative results on EnFoRe; (a)-(d) present cases where  EnFoRe correctly identifies the critical entities and retrieved question-relevant knowledge properly focuses on them; (e) and (f) present two failure cases.}
    \label{fig:qualitative_results}
\end{figure*}

\begin{table*}[t]
\small
\centering
\begin{tabular}{l|c|c|c|c}\toprule
\multirow{2}{*}{Sources} & \multicolumn{2}{c|}{RetTest} &  \multicolumn{2}{c}{RetTest Hard}\\\cline{2-5}
  &MRR@5 &  P@5  &   MRR@5   &  P@5   \\ \cline{1-5} 
 None  &  0.4632  & 0.3329      &  0.2525  &  0.1553 \\
 Image-based entities &  0.4688  & 0.3351      &   \textbf{0.2709}  &  \textbf{0.1637} \\
 Question-based entities &  0.4750  & 0.3409      &   0.2594  &  0.1612 \\
 Full   &\textbf{ 0.4800 }& 0.3444 & 0.2643 &  0.1632 \\\hline\hline
 w/o. Tags &   0.4788   &  0.3410  & 0.2624& 0.1606\\
 w/o. Wikidata &  0.4775  &  0.3429  &  0.2617& 0.1574\\
 w/o. Caption &   0.4794   &  \textbf{0.3449}  &  0.2626 & 0.1611\\ \hline
 w/o. Question & 0.4786   &  0.3442   & 0.2647 &0.1627\\
w/o. Sub-Question &  0.4784   &  0.3411 & 0.2625& 0.1605\\
w/o. Candidate  &   0.4693  &  0.3332 & 0.2664 & 0.1622\\ \bottomrule
\end{tabular}
\caption{Ablation study on entity sources.}
\label{tab:entity_source_ablation}
\end{table*}

We also present an ablation study on individual entity sources in Table \ref{tab:entity_source_ablation}. We introduce a particularly challenging ``RetTest Hard'' split that collects all of the examples in ``RetTest'' where none of the correct answers is in the entity set. Our EnFoRe model consistently achieves better retrieval performance (i.e. MRR@5 and P@5)  by incorporating entities extracted from each source. On the normal RetTest set, removing entities from candidate answers yields the largest decrease in MRR@5. This is due to the fact that the candidate answers cover plenty of correct answers in the OK-VQA test split and therefore provide direct hints to the desired content. On the RetTest Hard set, image-based entities generally help improve the retrieval performance more, indicating the need for explicitly discovering critical visual clues.

\subsection{Visual Question Answering Results}
We present the VQA performance of incorporating our EnFoRe knowledge in the state-of-the-art KAT model in Table \ref{tab:vqa_result}. While a plain KAT-base model, which uses GPT-3 and CLIP \cite{radford2021learning} to retrieve  image-based knowledge, achieves a score of (50.58)\footnote{The additional result shown in parentheses is computed by an unofficial evaluation metric 
that takes the max over 1.0 and number of annotators agreements divided by 3.
}, switching to our EnFoRe knowledge brings a 1.7 point improvements, achieving a score of 51.34 (52.24). Our ensemble model 
(KAT-full + EnFoRe) achieves a new SOTA score of 54.35 (55.23).

\noindent\textbf{Qualitative results:} We present sample results in Figure \ref{fig:qualitative_results} where (a)--(d) show cases where our EnFoRe model correctly identifies the critical entities ($i.e.$ the orange, the kite, the calico cat, and the teddy bear) and retrieved question-relevant knowledge focused on them. 
Case (e) shows an example where the retrieved sentence misleads the reader, because the reader currently only receives the textual input, and it fails to verify whether the pizza actually has a thin crust. Case (f) shows an example where the retriever properly focuses on the critical entity “NORWOOD” but fails to understand that this is the destination for the bus.

\noindent\textbf{Human evaluation:}
We also conducted a human evaluation on AMT of the retrieved entities and sentences to demonstrate that the knowledge retrieved by EnFoRe better supports the correct answers.  We first randomly sampled 1,000 test questions that are correctly answered by both the original KAT-base model and our “KAT-base + EnFoRe” model. Next, we extracted the top-$3$ sentences with the highest attention score averaged over all attention heads from the last decoder layer for both models. We also extracted the  top-3 visual entities. For EnFoRe, the top-$3$ entities with the highest attention scores in the input prompts are selected. For the original KAT model, we use the three entities from the three top retrieved sentences. Next, we show AMT workers the question, the predicted answer, the image with bounding boxes for the top entities, and the three retrieved sentences, for both systems randomly ordered. We present an example in the Appendix. Finally, workers are asked to judge which system's set of highlighted entities and sentences best supports the given answer. 
Experimental results show that judges pick our EnFoRe knowledge 61.8\% of the time, indicating a clear preference over the original KAT knowledge. Such information can be considered an explanation or rationale for the system's answer, and improved explanations can engender greater trust and acceptance from users and provide additional transparency of the system's operation.

\section{Conclusion}
In this work, we presented an Entity-Focused Retrieval (EnFoRe) model for retrieving knowledge for outside-knowledge visual questions. The goal is to retrieve question-relevant knowledge focused on critical entities. We first construct an entity set by parsing the question and the image. Then, EnFoRe  predicts a query-entity score, predicting how likely it will lead to finding a correct answer, and a passage-entity score showing how likely the entity fits in the context of the passage. These two scores are combined to re-rank the conventional query-passage relevancy score. EnFoRe demonstrates the clear advantages of improved multi-modal knowledge retrieval and helps improve VQA performance with its improved retrieved knowledge.
\section{Limitations}
Our EnFoRe model is empowered by a comprehensive set of parsed entities from the question and the image. However, as shown in the failure cases in the experiment section, those entities may contain detection errors that lead to undesired results. In addition, during training, we adopt a fully automatic scheme for annotating critical entities assuming they can help a sparse retriever achieve better SRR results; however, explicit human annotation could potentially improve the quality of the critical entities identified. 
While we have explored collecting both question-based and image-based entities in our current approach, they are not fully adequate in that ideally it could be beneficial to include not only the relevant objects for the visual question but other kinds of descriptors that may act as useful clues for knowledge retrieval. Another limitation of the current approach is that we encode each entity separately, ignoring the relationships between entities, which could be helpful for knowledge retrieval.

\section{Acknowledgements}
The research was supported by the NSF-funded Institute for Foundations of Machine Learning (IFML) at UT Austin. 
\bibliography{anthology,custom}
\bibliographystyle{acl_natbib}

\appendix

\section{Appendix}
\label{sec:appendix}
\subsection{Varying Weights during Re-Ranking}

\begin{figure}[h]
    \centering
    \includegraphics[clip, trim=1cm 16cm 23cm 0cm, width=\columnwidth]{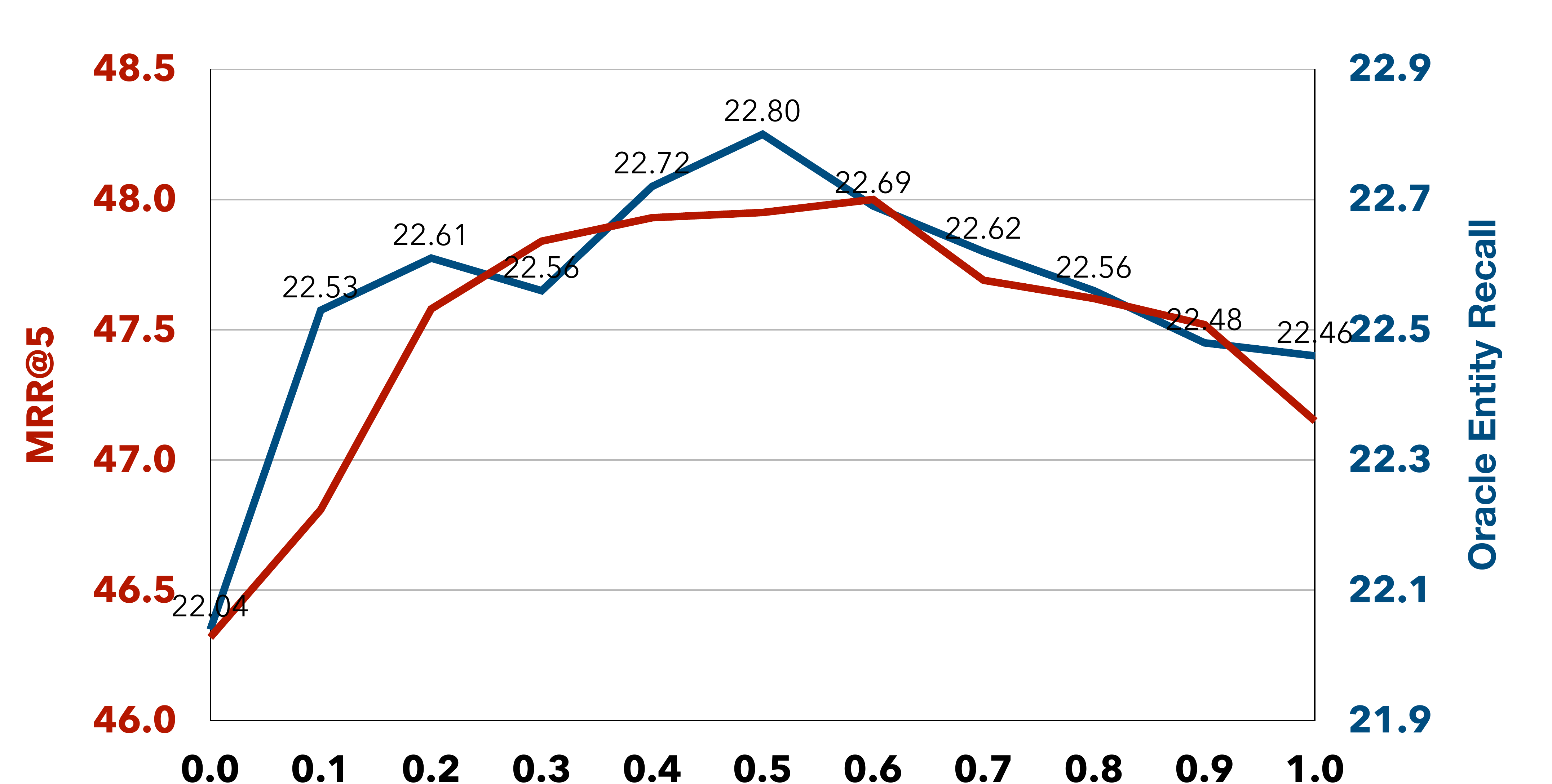}
    \caption{MRR and oracle-entity recall with different reranking weights.}
    \label{fig:weight}
\end{figure}

In Figure \ref{fig:weight}, we present the MRR (red line), and the oracle-entity recall (blue line) at a cut-off of 5, which is defined as the fraction of oracle entities appearing in the top-5 retrieved passages over the total number of the oracle entities. Our EnFoRe model not only improves the MRR results but also retrieves more oracle entities in the top passages, making the retrieved content more relevant. Also, the EnFoRe model is robust to the re-ranking weight, yielding consistent improvements for a broad range of weights.

\begin{figure*}[t]
    \centering
    \includegraphics[clip, trim=0cm 1cm 0cm 0cm, width=0.83\linewidth]{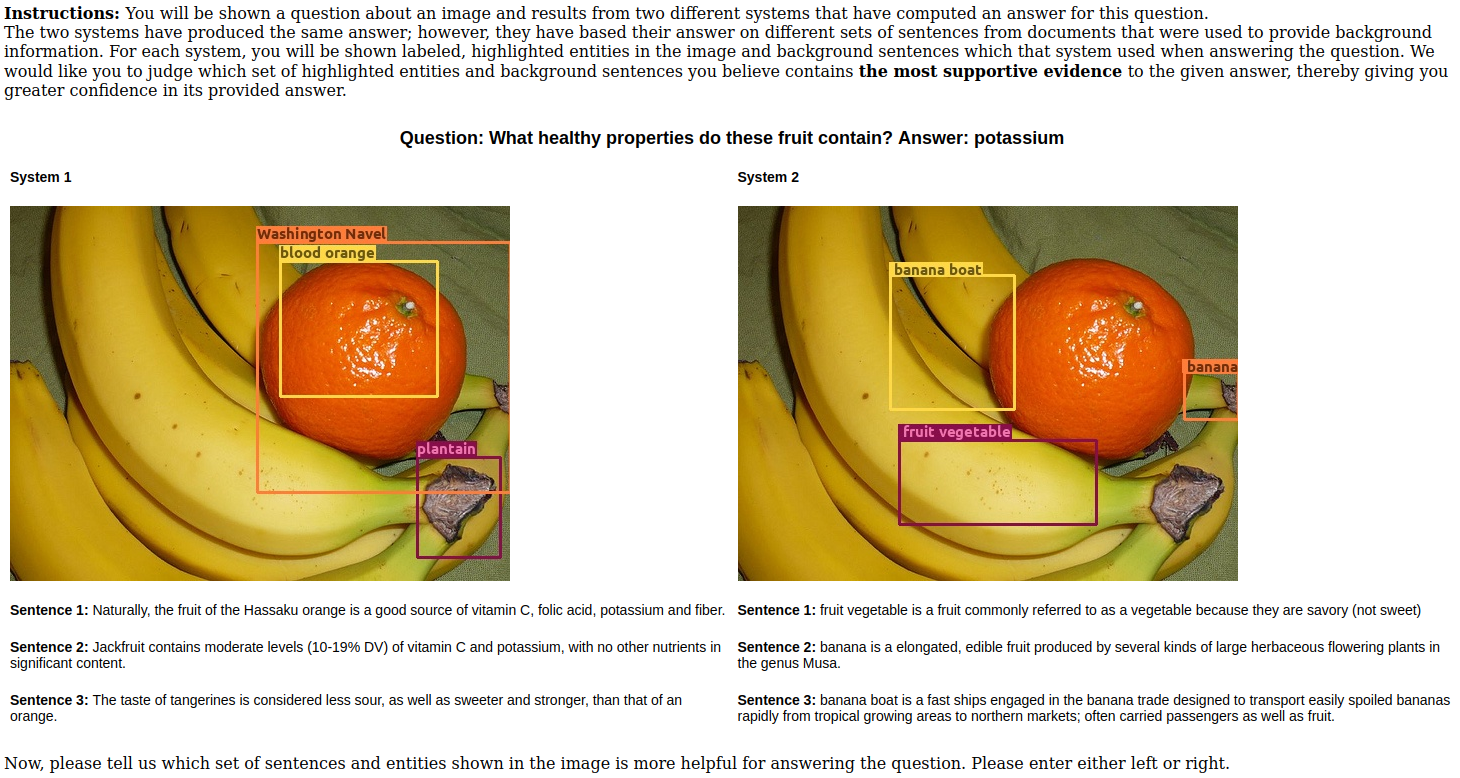}
    \caption{Sample question for the human evaluation.}
    \label{fig:human_eval_example}
\end{figure*}

\subsection{Human Evaluation Details}
We use Amazon Mechanical Turk (AMT) as our platform to perform human evaluation.  We randomly sample 1,000 test questions that are correctly answered by both the original KAT-base model and our “KAT-base + EnFoRe” model in order to focus on evaluating the explanations for their answers rather than their correctness. In each HIT (Human Inference Task), we include four questions together with a quality control example, where the preference should be clear. We eliminate data where the quality control is not passed, but pay the workers 80 cents for finishing the HIT regardless of passing the quality control example. The average time workers spent on each HIT is 2 min and 33 sec. Figure \ref{fig:human_eval_example} shows a sample question from a HIT.

\begin{table}[h]
\centering
\begin{tabular}{@{}l|l@{}}
\toprule
Hyperparameters           & Value     \\ \midrule
BM25 Retriever k          & 1.1       \\
BM25 Retriever b          & 0.4       \\
CLIP                      & ViT-B/16  \\ Learning rate             & 1e-5      \\
Optimizer                 & AdamW     \\
Batch size                & 6 per GPU \\
\#Gpus                    & 4         \\ 
Retriever hidden states   & 768       \\
Critical entity threshold & 0.8       \\
\#Epochs                  & 8         \\
Learning rate in KAT             & 3e-5\\
Optimizer  in KAT      & AdamW     \\
\#Sentences in KAT        & 10        \\
Batch size in KAT        & 24        \\
\bottomrule
\end{tabular}
\caption{Configurations for best-performing models.}
\label{tab:hyper}
\end{table}

\subsection{Hyperparameters}
We present details of the searching hyperparameters for the EnFoRe model in Table \ref{tab:hyper} . While most of the hyperparameters are set to the same as in \cite{qu2021passage}, we tune the threshold $\theta$ for recognizing critical entities (0.6, 0.8, 1.0), batch size (2, 4, 6 per GPU), and the number of training epochs (2, 4, 6). We use a greedy approach \cite{singh2018attention} to search hyperparameters in the order of $\theta$, batch size, and training epochs. Maximizing MRR@5 is used as the objective.

\end{document}